%% file: main.tex
\documentclass[10pt,twocolumn,letterpaper]{article}

\usepackage[final]{cvpr}      

\usepackage{dsfont}
\usepackage{amsmath,amssymb}
\usepackage{float}

\input{preamble}

\definecolor{cvprblue}{rgb}{0.21,0.49,0.74}
\usepackage[pagebackref,breaklinks,colorlinks,citecolor=cvprblue]{hyperref}

\def\paperID{{8768}} 
\def\confName{CVPR}
\def\confYear{2025}

\input{Figure/fig1}

\title{Toward Scene Graph and Layout Guided Complex 3D Scene Generation}

\author{Yu-Hsiang Huang$^1$\textcolor{white}, \quad Wei Wang$^{1}$, \quad 
Sheng-Yu Huang$^{1}$, \quad Yu-Chiang Frank Wang$^{1, 2}$, \\
\normalsize \textsuperscript{1} National Taiwan University \quad \textsuperscript{2} NVIDIA, Taiwan\\
{\tt\small \{b09901062, b09901015, f08942095\}@ntu.edu.tw frankwang@nvidia.com}
}

\begin{document}

\def\frank{\textcolor{red}}
\def\pj{\textcolor{green}}
\def\kids{\textcolor{blue}}
\newcommand{\tabincell}[2]{\begin{tabular}{@{}#1@{}}#2\end{tabular}}
\newlength\savewidth\newcommand\shline{\noalign{\global\savewidth\arrayrulewidth\global\arrayrulewidth1.25pt}\hline\noalign{\global\arrayrulewidth\savewidth}}
\maketitle

\input{sec/0_abstract}

\input{sec/1_intro}
\input{Figure/fig2}
\input{sec/2_relate}

\input{sec/3_method}
\input{sec/4_experiments}

\input{sec/5_conclusion}
{    \small
    \bibliographystyle{ieeenat_fullname}
    \bibliography{main}
}
\input{sec/X_suppl}

\end{document}

%% file: preamble.tex
\usepackage[dvipsnames]{xcolor}

\usepackage[accsupp]{axessibility}
\usepackage[normalem]{ulem}
\usepackage{multirow}
\usepackage{colortbl}
\usepackage{pifont}
\usepackage{bbding}
\def\halfcheckmark{\ding{52}\textsuperscript{\kern-0.50em\small\ding{55}}}

\newcommand{\frank}[1]{\textcolor{red}{#1}}

%% file: Figure/fig1.tex
\makeatletter
\g@addto@macro\@maketitle{
\begin{figure}[H]
	\setlength{\linewidth}{\textwidth}
        \setlength{\hsize}{\textwidth}
        \centering

        \vspace{-10mm}

	\includegraphics[width=0.8\textwidth]{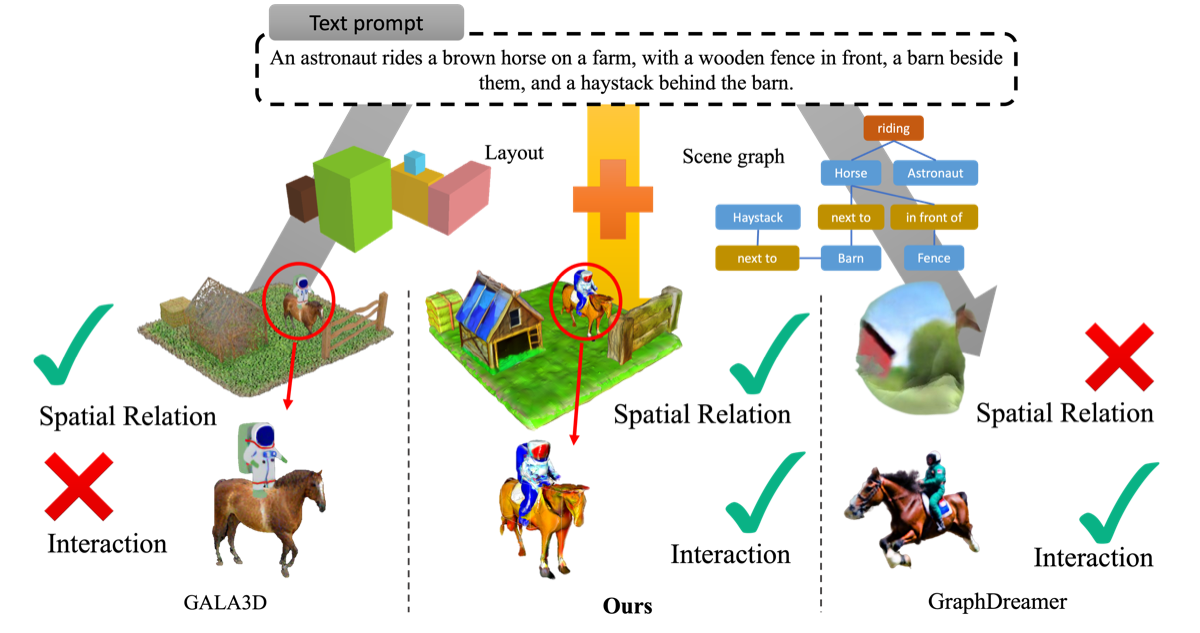}
    \caption{Illustration of complex 3D scene generation. Existing text-to-3D methods generally model the layout between objects (e.g., GALA3D~\cite{Zhou2024GALA3DTT}) or focus on object-centric generation with interaction observed (e.g., GraphDreamer~\cite{Gao2023GraphDreamerC3}). Our GraLa3D aims to perform complex 3D scene generation, with both spatial relation and interaction between objects properly preserved.}
    
	\label{fig:archi}
\end{figure}
}
\makeatother

%% file: sec/0_abstract.tex
\begin{abstract}
Recent advancements in object-centric text-to-3D generation have shown impressive results. However, generating complex 3D scenes remains an open challenge due to the intricate relations between objects. Moreover, existing methods are largely based on score distillation sampling (SDS), which constrains the ability to manipulate multi-objects with specific interactions. Addressing these critical yet underexplored issues, we present a novel framework of Scene \textbf{Gra}ph and \textbf{La}yout Guided \textbf{3D} Scene Generation (GraLa3D). Given a text prompt describing a complex 3D scene, GraLa3D utilizes LLM to model the scene using a scene graph representation with layout bounding box information. GraLa3D uniquely constructs the scene graph with single-object nodes and composite super-nodes. In addition to constraining 3D generation within the desirable layout, a major contribution lies in the modeling of interactions between objects in a super-node, while alleviating appearance leakage across objects within such nodes. Our experiments confirm that GraLa3D overcomes the above limitations and generates complex 3D scenes closely aligned with text prompts.
\end{abstract}

%% file: sec/1_intro.tex
\section{Introduction}
\label{sec:intro}

The creation of intricate 3D assets has traditionally been a labor-intensive process, requiring considerable time and specialized skills from designers. Recent advances in generative artificial intelligence \cite{Metzer2022LatentNeRFFS, Wang2022ScoreJC, Poole2022DreamFusionTU, Xu2022Dream3DZT} have significantly transformed this landscape by enabling the synthesis of 3D content directly from natural language descriptions. These advancements can be broadly categorized into two domains: object-centric and scene-level text-to-3D generation. For instance, an object-centric prompt such as “a 3D model of a vintage wooden chair" results in a standalone model of a chair. In comparison, a scene-level prompt like “a cozy living room with a vintage wooden chair, a coffee table, and a bookshelf by the window" aims to generate an entire scene where multiple objects are properly arranged. While object-centric generation focuses on individual assets, scene-level generation plays a crucial role in applications such as gaming, virtual reality, and autonomous simulation, where the synthesis of coherent, contextually rich environments is indispensable~\cite{Li2023GenerativeAM}.

Recent methods of text-to-3D generation leverage 2D priors derived from large pre-trained text-to-image 
 diffusion models~\cite{Rombach2021HighResolutionIS, Saharia2022PhotorealisticTD}. Specifically, DreamFusion~\cite{Poole2022DreamFusionTU} introduces the technique of Score Distillation Sampling (SDS) loss. Given a text prompt, a randomly initialized 3D representation is rendered into multiple 2D views, which are then used as input to a pre-trained text-to-image diffusion model. The denoising results from these 2D renderings are directly utilized to optimize the 3D representation, ensuring that the generated 3D asset aligns visually with the input text prompt. Building on the concept of SDS loss, subsequent efforts further enhance the quality and coherence of generated assets by employing a coarse-to-fine strategy to improve resolution \cite{Lin2022Magic3DHT}, utilizing multi-view diffusion models to ensure consistency across different angles \cite{Shi2023MVDreamMD} or improving the SDS loss itself to provide a more precise guidance~\cite{Liang2023LucidDreamerTH}. While these methods represent significant strides in object-centric 3D generation, they face substantial limitations in adapting to scene-level generation tasks due to the inherent limitations of diffusion models in handling multi-object compositions, as noted in \cite{Zhang2023ASO, Ma2023DirectedDD}.

To effectively address the complexities of scene-level text-to-3D generation, one stream of works \cite{Lin2023CompoNeRFTM, cohenbar2023setthescenegloballocaltraininggenerating, Po2023Compositional3S, wang2024luciddreaming} decomposes the scene into a composition of individual 3D objects while each object is generated separately with given layouts. Among them, GALA3D~\cite{Zhou2024GALA3DTT}  proposes utilizing large language models (LLMs, e.g., GPT-3.5~\cite{openai2024gpt4technicalreport} to generate non-overlapping layouts from text prompts automatically. This approach generates scenes with a clear spatial layout but works well only for scenes with non-interacting objects, as non-overlapping layouts cannot effectively model interactions between objects. To address this, GraphDreamer~\cite{Gao2023GraphDreamerC3} proposed another research direction employing scene graphs, where objects (nodes) and their relations (edges) are used to represent complex interactions within the scene. This graph-based structure allows for iterative optimization of each object and the relations between them, providing the scene with vivid interactions. However, it faces challenges in accurately modeling the spatial relations between elements, often leading to subtle inconsistencies in layout arrangement due to the inherent limitations of the underlying diffusion models as noted in~\cite{wen2023improvingcompositionaltexttoimagegeneration}.

In this paper, we propose a Scene \textbf{Gra}ph and \textbf{La}yout Guided \textbf{3D} Scene Generation (GraLa3D), which combines the strengths of both compositional layouts and scene graphs while addressing the limitations of existing approaches. Given a text prompt describing a complex 3D scene, both a scene graph and layout bounding boxes are generated by the LLM. Specifically, our GraLa3D models the scene graph via single-object nodes and supernodes with our Scene Graph Composition stage, where spatial relations and interactions between each object are handled, respectively. Along with the constraint provided by the layout bounding boxes to model the spatial relations, our GraLa3D further generates objects with interaction relations properly addressed in supernodes in our Node-to-3D Generation stage, while preventing disentanglement across different objects, achieving ideal text-to-complex-3D-scene generation.

The contributions of our work are as follows:
\begin{itemize}

\item We propose GraLa3D, a text-to-3D generation framework that utilizes scene graph and layout information to generate complex 3D scenes so that spatial and interaction relationships between objects can be properly modeled.
\item With our Scene Graph Composition stage, we construct the scene graph as single-object nodes and supernodes, where each model objects that are involved with spatial relations solely and objects involved in interaction relations, respectively.
\item Our Node-to-3D Generation stage produces objects described by single-object nodes and supernodes, not only generating objects with proper size and location described by the corresponding layouts but also providing vivid interactions while alleviating undesirable entanglement between objects.
\item By rearranging all objects spatially according to the layout bounding boxes, the final 3D Scene Harmonization stage fuses all the objects together with global style guidance, ensuring the coherent appearance of the entire 3D scene.
\end{itemize}



%% file: Figure/fig2.tex
\begin{figure*}[ht]
    \centering

        \includegraphics[width=1\linewidth]{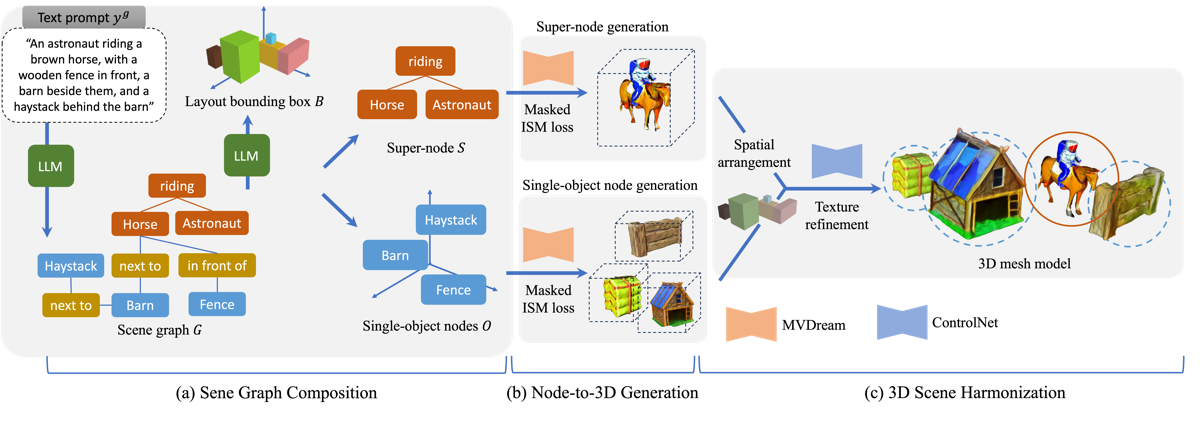}

    \vspace{-4mm}
    \caption{\textbf{Method Overview of GraLa3D.} The proposed method consists of three stages: (a) Scene Graph Composition, (b) Node-to-3D Generation, and (c) 3D Scene Harmonization. Stage (a) converts a text prompt \( y^g \) into a scene graph \( G \) representation with the associated layout bounding boxes \( B \). With nodes corresponding to objects with or without interaction, Stage (b) generates 3DGS aligned with the information described in (a). Finally, Stage (c) enforces the output scene to exhibit proper appearance and texture consistency.
}
    \label{fig:pipeline}
\end{figure*}

%% file: sec/2_relate.tex
\section{Related Work}
\label{sec:related}

\subsection{Object-Centric Text-to-3D Generation}

Early methods of object-centric text-to-3D generation~\cite{Sanghi2021CLIPForgeTZ,Chen2018Text2ShapeGS,Fu2022ShapeCrafterAR,Liu2022TowardsIT} primarily rely on pre-existing 3D asset datasets like ShapeNet~\cite{Chang2015ShapeNetAI} for training. However, the scarcity of paired text-shape data limits the generalization ability of these models. To address the dependence on paired datasets, subsequent approaches such as DreamFields~\cite{Jain2021ZeroShotTO}  and CLIP-Mesh~\cite{Khalid2022CLIPMeshGT} leverage pre-trained vision-language models (VLMs)~\cite{Radford2021LearningTV} to align 2D renderings of 3D objects with text prompts. These methods are constrained, however, by the VLMs' limited understanding of complex textual queries and 3D structures, often resulting in suboptimal fidelity and detail in the generated objects~\cite{Sanghi2022CLIPSculptorZG}.

The introduction of diffusion models~\cite{Rombach2021HighResolutionIS, Saharia2022PhotorealisticTD}, marks a significant step forward for text-to-3D generation. DreamFusion~\cite{Poole2022DreamFusionTU} introduces the Score Distillation Sampling (SDS) loss. In this approach, a randomly initialized 3D representation is rendered into multiple 2D views, which are then fused with noise and processed by a 2D diffusion model for denoising. The difference between the predicted and injected noise is used to directly optimize the parameters of the 3D representation. This innovation allows text-to-3D generation to move beyond reliance on existing 3D shape datasets. Nevertheless, there are limitations like the low resolution in generated objects, and the tendency of these models to focus primarily on front-facing views often leads to 3D inconsistencies (commonly referred to as the "Janus problem"~\cite{Armandpour2023ReimagineTN, Shi2023MVDreamMD}). 

To alleviate these issues, subsequent researches introduce refined optimization processes or more effective diffusion guidance. For example, MVDream~\cite{Shi2023MVDreamMD}  improves multiview consistency by finetuning the diffusion model on a 3D dataset with an additional multiview self-attention module. LucidDreamer~\cite{Liang2023LucidDreamerTH} further enhances the reliability and consistency of 2D diffusion guidance by proposing Interval Score Matching, which utilized deterministic diffusing trajectories~\cite{song2022denoisingdiffusionimplicitmodels} and interval-based score matching~\cite{Liang2023LucidDreamerTH} to mitigate over-smoothing. Although these methods achieve success in object-centric text-to-3D generation, they encounter difficulties when attempting to generate complex 3D scenes, as diffusion models generally perform suboptimally when handling input prompts involving multiple objects or spatial relationships between them~\cite{Zhang2023ASO, Ma2023DirectedDD}.

\subsection{Scene-Level Text-to-3D Generation}

To better model scene-level text-to-3D generation, one mainstream solution is to compositionally generate the scene by breaking the scene into separated parts and generating the parts individually. For instance, CompoNeRF~\cite{Lin2023CompoNeRFTM} utilizes handcrafted layouts consisting of a bounding box and the spatial position of each object as constraints to first generate objects separately and then arrange the generated objects according to their spatial position. GALA3D~\cite{Zhou2024GALA3DTT} extends this idea by leveraging LLM-generated non-overlapping layout bounding boxes as the initial constraint to mitigate manual labor. Each object is then generated individually with guidance from instance-level diffusion models \cite{Shi2023MVDreamMD} and scene-level diffusion priors \cite{zhang2023adding} to better reflect the text prompt. While effective for scenes with simple spatial configurations, the non-overlapping layout assumption limits its applicability when objects should contact each other according to the description (e.g., riding, holding), making it challenging to represent complex scenes with intricate object interactions.

To overcome the limitation of layout-guided 3D scene generation, GraphDreamer~\cite{Gao2023GraphDreamerC3} proposes a framework using scene graphs to describe complex relationships between objects. A scene graph consisting of objects (nodes) and relationships (edges) is first generated from the text prompt. Then, to produce results reflecting the graph, paired objects and their interactions are iteratively optimized using node-wise and edge-wise SDS loss. While this method improves the handling of object interactions, GraphDreamer still encounters difficulties in optimizing spatial relationships due to the inherent limitations of diffusion models in comprehending spatial arrangements as described in~\cite{wen2023improvingcompositionaltexttoimagegeneration}. Additionally, as the number of objects in the scene increases, scalability becomes a pressing issue due to the possibility of exponential growth in the number of relationships to be managed. These challenges highlight the need for an approach that can effectively model both complex spatial arrangements and interactions while maintaining scalability in scene-level generation.

%% file: sec/3_method.tex
\section{Method}
\subsection{Problem Formulation and Model Overview}
As depicted in Figure~\ref{fig:pipeline}, we propose a text-to-3D complex-scene generation scheme, conditioned on a text prompt $y^g$ which describes a scene that involves spatial and interactive relations between $n$ different objects. In order to accurately reflects both types of relationships described in $y^g$, our framework is composed of three stages: \textit{Scene Graph Composition, Node-to-3D Generation, and 3D Scene Harmonization}. 

For the stage of Scene Graph Composition, an LLM~\cite{openai2024gpt4technicalreport} is utilized to convert $y^g$ into a scene graph description $G = (V, E)$,  where \( V = \{ v_i \}_{i=1}^n \) represents the \( n \) objects (nodes) described in $y^g$ and \( E = \{ e_{i,j} \} \) represents the relations (edges) between them. To better describe objects with and without interactions, we uniquely decompose $V$ into single-object nodes $O$ and super-nodes $S$, with the associated bounding boxes \( B = \{ b_i \}_{i=1}^n \) described by the LLM. We advance 3D Gaussian Splatting (3DGS) in the stage Node-to-3D Generation, aiming to produce 3DGS models that accurately aligns with $O$ and $S$. Finally, 3D Scene Harmonization maps the updated 3DGS into 3D meshes, optimizing the textures with appearance and style consistency.

\subsection{Scene Graph Composition}
\input{Figure/fig3}

As the first stage of our proposed framework, we leverage the LLM~\cite{openai2024gpt4technicalreport} to convert the input text prompt $y^g$ into a structured scene graph $G$ and bounding boxes $B$. With nodes and edges in $G$ describe the objects and their relationships in $y^g$, $B$ denotes the spatial information for each object, and thus suggest the 3D layout (refer to our supplementary material for this process).

We note that, based on the different relation types described in $G$ (i.e., \textit{spatial relation} vs. \textit{interaction}), we particularly have LLM decompose the nodes $V$ of $G$ into two groups: single-object nodes $O$ and super-nodes $S$. Take Figure~\ref{fig:decomp} as an example, single-object nodes $O = \{ v^o_1, v^o_2, v^o_3 \}$ denote the objects that are only connected by spatial relations (i.e., haystack, barn, and fence), while super-nodes $S = \{v^S_1, v^S_2, e^S_{1,2} \}$ contain the objects with interactive relations (e.g., \textit{astronaut-horse} with \textit{riding}). Our key insight for super-node construction is based on the observation that only interactions between objects need to be optimized with a diffusion prior, thus modeling such object-relation-object triplets with a super-node~\cite{Gao2023GraphDreamerC3} benefits the distillation of knowledge from text-to-image diffusion models. On the other hand, spatial relations between objects can be easily defined by bounding boxes \( B \) and do not heavily depend on the diffusion models as noted in~\cite{Zhou2024GALA3DTT}. 


\subsection{Node-to-3D Generation}
\label{sec:nodegen}
\paragraph{3D generation from a single node.} \label{sec:singlenode}
To generate 3D models for each \( v^o_i \) within the corresponding bounding box (denoted as $ b^o_i$), a 3DGS model $\theta^o_i$ is initialized within the spatial area described by $b^o_i$. Then, we adopt the approach of leveraging a 2D diffusion prior~\cite{Shi2023MVDreamMD} as guidance with object name $y^o_i$ of $v^o_i$ as a condition for the 3D generation, in line with established methods~\cite{tang2024dreamgaussian,Poole2022DreamFusionTU}. Specifically, MVDream~\cite{Shi2023MVDreamMD} is employed as our multi-view consistent diffusion prior $\epsilon_\phi$, and Interval Score Matching (ISM)~\cite{Liang2023LucidDreamerTH} is adopted as a guiding framework. 

However, without any spatial constraints, this object-centric generation method might lead 3D models to extend beyond the predicted layout boundaries $b^o_{i}$, reducing spatial accuracy with respect to the scene prompt. To address this, we propose to enforce the alignment the generated 3DGS within the designated scene structure. Specifically, an explicit 2D layout constraint is formulated to punish Gaussian blobs that extend outside the mask:
\begin{equation}
\mathcal{L}_{\text{layout}}(\theta^o_{i}) = \left\lVert \alpha^o_i \cdot (\mathds{1} - Proj(b^{o}_{i}, c)) \right\rVert_1,
\end{equation}
where $\alpha^o_i$ represents the rendered alpha image of $\theta^o_i$ and $Proj(b^{o}_{i}, c)$ is the projected 2D bounding box given a camera pose $c$. 

Although the above $\mathcal{L}_{\text{layout}}$ ensures that the generated object lies inside its bounding box, we observe that the 2D diffusion prior tends to generate an incomplete and oversized object that fits the bounding box instead of a complete object with a proper size. Inspired by~\cite{repaint2022}, we introduce a masked ISM loss to alleviate this problem, further constraining the 2D diffusion prior as guidance within the layout of interest. Thus, our ISM loss is computed based on sampled noise only within $Proj(b^{o}_{i}, c)$, i.e.,
\begin{equation}
\mathcal{L}^{\text{mask}}_{\text{ISM}}(\theta^o_{i},y^o_i) = \mathcal{L}_{\text{ISM}} \cdot Proj(b^{o}_{i}, c) ,
\end{equation}
where $\mathcal{L}_{\text{ISM}(\cdot)} $ represents the original ISM loss~\cite{Liang2023LucidDreamerTH} to guide the generation of $\theta^o_{i}$ conditioned on object prompt $y^o_i$ .
Consequently, the overall objective function for single-node generation combines these elements, calculated as:
\begin{equation}
\mathcal{L}_{\text{single}}(\theta^o_{i},y^o_i) = \mathcal{L}^{\text{mask}}_{\text{ISM}}(\theta^o_{i},y^o_i) + \mathcal{L}_{\text{layout}}(\theta^o_{i}).
\end{equation}

\paragraph{3D generation from a super-node.} \label{sec:super-node}
\input{Figure/fig4}
To generate 3DGS for each super-node $S = \{v^S_i, v^S_j, e^S_{i,j} \}$ with the text prompt $y^S_{i,j}$ (e.g., ``an astronaut riding a horse''), we need to ensure the interaction between $\{v^S_i, v^S_j\}$ is properly modeled in within the corresponding bounding boxes $\{b^S_i, b^S_j\}$. Figure~\ref{fig:supernode} gives a detailed illustration of this process, where 3DGS models $\theta^S_{i}$ and $\theta^S_{j}$ are initialized for objects $v^S_i$ and $v^S_j$ within bounding boxes $b^S_{i}$ and $b^S_{j}$, respectively. By separately initializing $\theta^S_{i}$ and $\theta^S_{j}$, a relation branch needs to be additionally constructed, jointly optimizing $\{v^S_i, v^S_j\}$ while aligned with their interaction description. And, for the object instance generation, the same approach is applied as that for single nodes.

For the above relation branch, by giving the relation prompt $y^S_{i,j}$ and a randomly sampled camera pose $c$, an interaction loss $\mathcal{L}_{\text{int}}$ is applied to the union of $\theta^S_i$ and $\theta^S_j$ (i.e., $\theta^S_i \cup \theta^S_j$) as:
\begin{equation}
    \begin{aligned}
    \mathcal{L}_{\text{int}}(\theta^S_{1} \cup \theta^S_{2},y^S_{1,2}) &= \mathcal{L}^{\text{mask}}_{\text{ISM}}(\theta^S_{1} \cup \theta^S_{2},y^S_{1,2}) \\
    & + \mathcal{L}_{\text{layout}}(\theta^S_{1} \cup \theta^S_{2}),
    \end{aligned}
\end{equation}
where the 2D mask required for both losses is projected from the union bounding box $b^S_{1} \cup b^S_{2}$ and the masked ISM loss $\mathcal{L}^{\text{masked}}_{\text{ISM}}$ is applied by using $y^S_{1,2}$ as input. 

\input{Table/main_table}
\input{Figure/4obj_qualitative}

However, applying the interaction loss to $\theta^S_i \cup \theta^S_j$ might result in certain parts (and their corresponding Gaussians) from $v^S_1$ erroneously categorized as that of $v^S_2$, and vice versa. To further tackle this problem, we introduce a localization loss  $\mathcal{L}_{\text{local}}$, enforcing the alignment between the 2D mask of $\theta^S$ and its attention map $D^S$. This provides a necessary instance-level constraint from the attention mechanism to localize a proper location and silhouette for $v^S_2$, in the presence of the joint optimization of $\theta^S_i \cup \theta^S_j$. The attention map, derived from $x^S_{i,j}$ for the prompt $y^S_j$, highlights the region associated with $v^S_j$ and excludes $v^S_i$. This alignment guides $\theta^S_2$ to represent the correct object, $v^S_2$, instead of $v^S_1$. We thus define the localization loss as follows:
\[
\mathcal{L}_{\text{local}}(\theta^S_{2}) = \left\lVert \alpha^S_{2} - D^S_{2} \right\rVert_2^2,
\]
where \(\alpha^S_2\) is the rendered alpha mask for the 3D model \(\theta^S_2\), 
and \(D^S_2\) is the cross-attention map that marks the region in the rendered image \(x^S_{1,2}\) related to the token of \(v^S_2\) in the prompt $y^S_{1,2}$.  Therefore, for the object branch, the object loss $\mathcal{L}_{obj}$ for $v^S_2$ is formulated as:

\begin{equation}
\begin{aligned}
    \mathcal{L}_{\text{obj}}(\theta^S_{2},y^S_2) &= \mathcal{L}^{\text{mask}}_{\text{ISM}}(\theta^S_{2},y^S_2) + \mathcal{L}_{\text{layout}}(\theta^S_{2}) \\
    &+ \mathcal{L}_{\text{local}}(\theta^S_{2}), 
    \end{aligned}
\end{equation} 
where $y^S_2$ represents the text prompt ``horse''. Similarly, $\mathcal{L}_{obj}$ for $v^S_1$ is calculated as:
\begin{equation}
\begin{aligned}
    \mathcal{L}_{\text{obj}}(\theta^S_{1},y^S_1) &= \mathcal{L}^{\text{mask}}_{\text{ISM}}(\theta^S_{1},y^S_1) + \mathcal{L}_{\text{layout}}(\theta^S_{1}) \\
    &+ \mathcal{L}_{\text{local}}(\theta^S_{1}).
    \end{aligned}
\end{equation} 

The overall objective for 3D upernode generation is defined as: 
\begin{equation}
\begin{aligned}
    \mathcal{L}_{\text{super}}(\theta^S_{1} \cup \theta^S_{2},y^S_{1,2}) & = \mathcal{L}_{\text{int}}(\theta^S_{1} \cup \theta^S_{2},y^S_{1,2}) + \mathcal{L}_{\text{obj}}(\theta^S_{1},y^S_{1}) \\
    & + \mathcal{L}_{\text{obj}}(\theta^S_{2},y^S_{2}). 
\end{aligned}
\end{equation} 

\noindent To sum up, with $\mathcal{L}_{single}$ for each single-object node and $\mathcal{L}_{super}$ for the super-node, all the objects described by $O$ and $S$ can be generated via 3DGS accordingly.
\label{sec:scenedecomposition}

\subsection{3D Scene Harmonization}
 As the final stage of 3D scene generation, we need to ensure the 3D objects produced by distinct 3DGS exhibit visual consistency in the scene of interest. Therefore, we present a Global Style Harmonization strategy to refine the texture of all generated objects together with the global prompt $y^g$. 
 
 Inspired by DreamGaussian~\cite{tang2024dreamgaussian}, we extract mesh models for all optimized Gaussians (each mesh model corresponds to an object) and only refine the UV maps $\mathcal{M}_{UV}$ of the mesh models to prevent undesirable modifications on the geometry. The guidance for the refinement is a pre-trained depth-conditioned ControlNet~\cite{zhang2023adding} and the objective is formulated as the pixel-wise MSE loss as:

\begin{equation}
    \mathcal{L}_{\text{MSE}}(\mathcal{M}_{UV}) = \left\| x^g - x^{\text{refine}} \right\|_2^2,
\end{equation}
where \( x^g \) is the rendered global image, and \( x^{\text{refine}} \) is obtained by perturbing \( x^g \) with random noise and applying a multi-step denoising process using the ControlNet with $y^g$ as the input. To this end, we are able to integrate the aforementioned different generation approaches and produce a cohesive 3D mesh scene, achieving a clear layout and enhanced interactivity.

 \input{Figure/5obj_qualitative}
\input{Figure/kitchen_qualitative}


%% file: Figure/fig3.tex
\begin{figure}[t]
    \centering
    \includegraphics[width=\columnwidth]{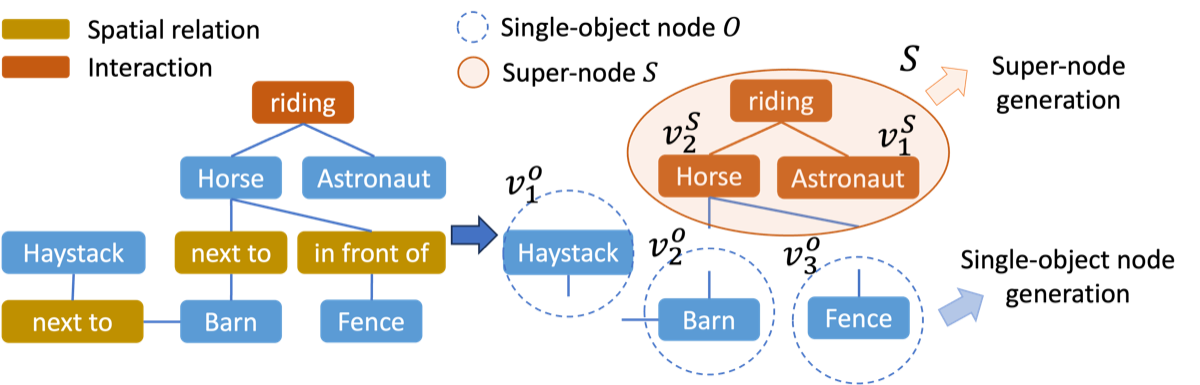}
    \vspace{-2mm}
    \caption{\textbf{Scene graph composition. } We utilize LLM to construct a scene graph describing objects and their relations. In particular, nodes in blue denotes single objects in the scene, while supernodes in orange describe objects with interactions.}
    \label{fig:decomp}
\end{figure}

%% file: Figure/fig4.tex
\begin{figure}[t]
    \centering
    \includegraphics[width=\columnwidth]{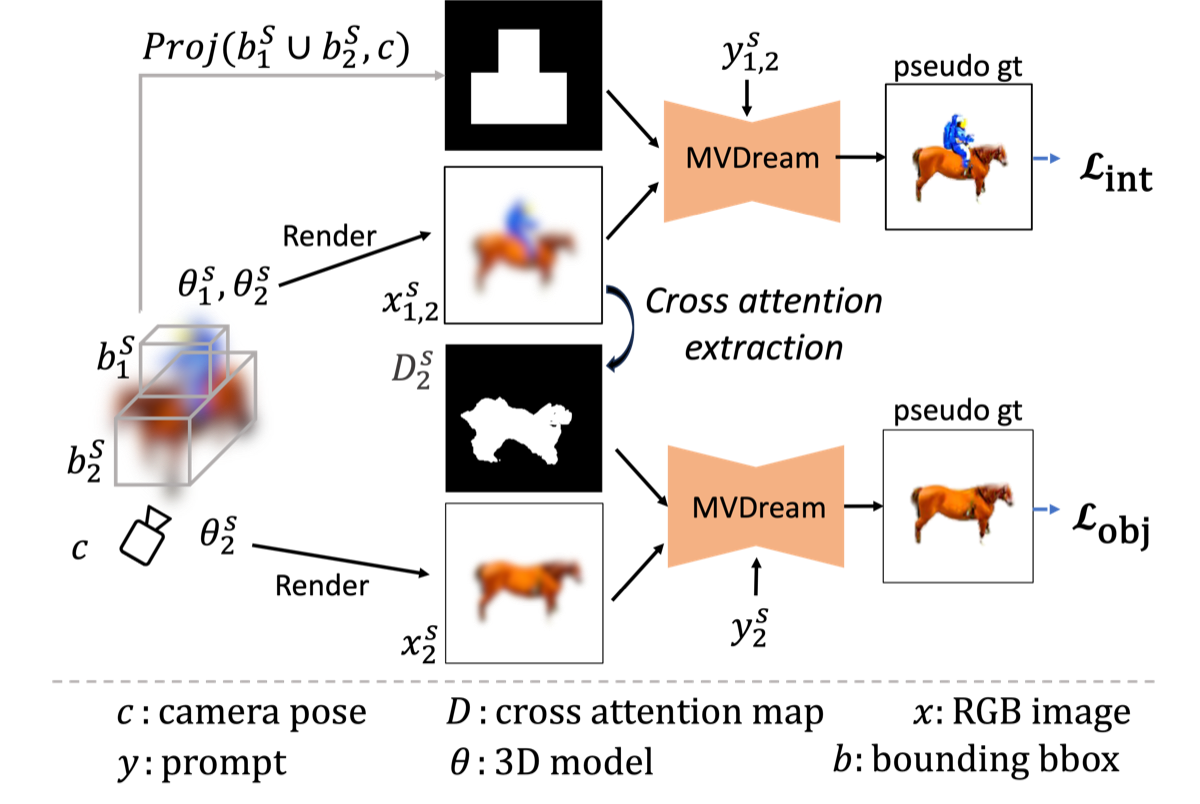}
    \vspace{-2mm}
    \caption{\textbf{Super-node Generation.} Given $y^S_{1,2}$ as input, 3DGS models $\theta^S_1$, $\theta^S_2$ are initialized and optimized. In the upper branch, $\theta^S_1$, $\theta^S_2$ are jointly optimized using $y^S_{1,2}$ and $b^S_1 \cup  b^S_2$. 
    In the lower branch, taking the horse ($y^S_2$) as an example, the attention map $D^S_2$, which corresponds to the token of $y^S_2$ from $x^S_{1,2}$, is used as guidance to localize the generation region of $x^S_2$.} 
    \label{fig:supernode}
\end{figure}

%% file: Table/main_table.tex
\begin{table*}[t]
\centering
 \caption{\textbf{Quantitative evaluation in terms of CLIP scores.} We report average CLIP scores using different prompts and 200 images rendered at random views. Please refer to each figure example for the text prompts used. }
 \label{tab:main_table}
\resizebox{\textwidth}{!}{%
\begin{tabular}{l|c|c|c|c|c|c| c|c}
                         & Representation & Farm case (Fig.~\ref{fig:archi}) &Wizard case (Fig.~\ref{fig:4obj_qualitative}) &Kitchen case (Fig.~\ref{fig:kitchen_qualitative}) & Mermaid case(Fig.~\ref{fig:5obj_qualitative})& Bear case (Fig.~\ref{fig:5obj_qualitative}) & Avg. \\ \hline
GraphDreamer~\cite{Gao2023GraphDreamerC3}&SDF& 0.252 & 0.279 & 0.283 & 0.236 & 0.115  & 0.233\\
GALA3D~\cite{Zhou2024GALA3DTT}&3DGS& 0.283& 0.264&0.275& 0.314 & 0.268& 0.281\\
GraLa3D (Ours)& 3DGS $\rightarrow$ Mesh& 0.336& 0.311&0.293& 0.321& 0.277 &0.308\\
    \vspace{0.5em}
\end{tabular}}
\end{table*}

%% file: Figure/4obj_qualitative.tex
\begin{figure*}[t]
    \centering

        \includegraphics[width=1\linewidth]{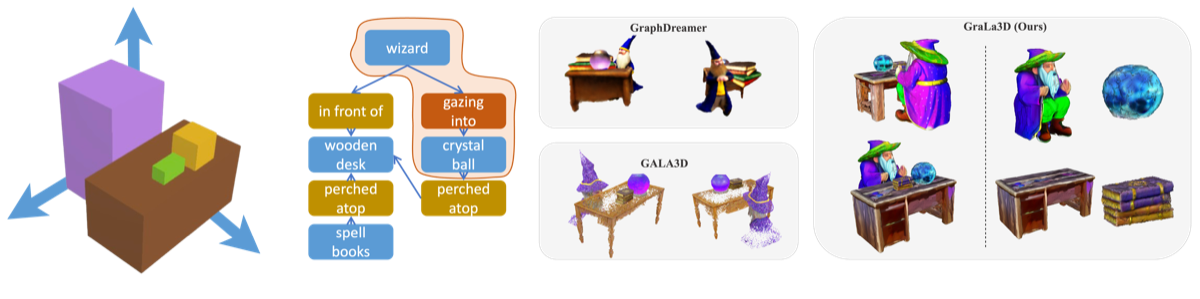}

    \vspace{-2mm}
    \caption{\textbf{Example text-to-3D generation with four objects.} Given the text prompt of ``\textit{a Wizard in front of a Wooden Desk, gazing into a Crystal Ball perched atop the Wooden Desk, with a Stack of Ancient Spell Books perched atop the Wooden Desk}'', we are able to generate a wizard-crustal ball pair with proper spatial and interaction relationship, with the ball and books placed on the table. On the other hand, GALA3D fails to generate the interaction between the wizard and the crystal ball, while GraphDreamer fails to produce a proper 3D layout (e.g., the books are larger than the table).
}
    \label{fig:4obj_qualitative}
\end{figure*}

%% file: Figure/5obj_qualitative.tex
\begin{figure*}[t!]
    \centering

        \includegraphics[width=1\linewidth]{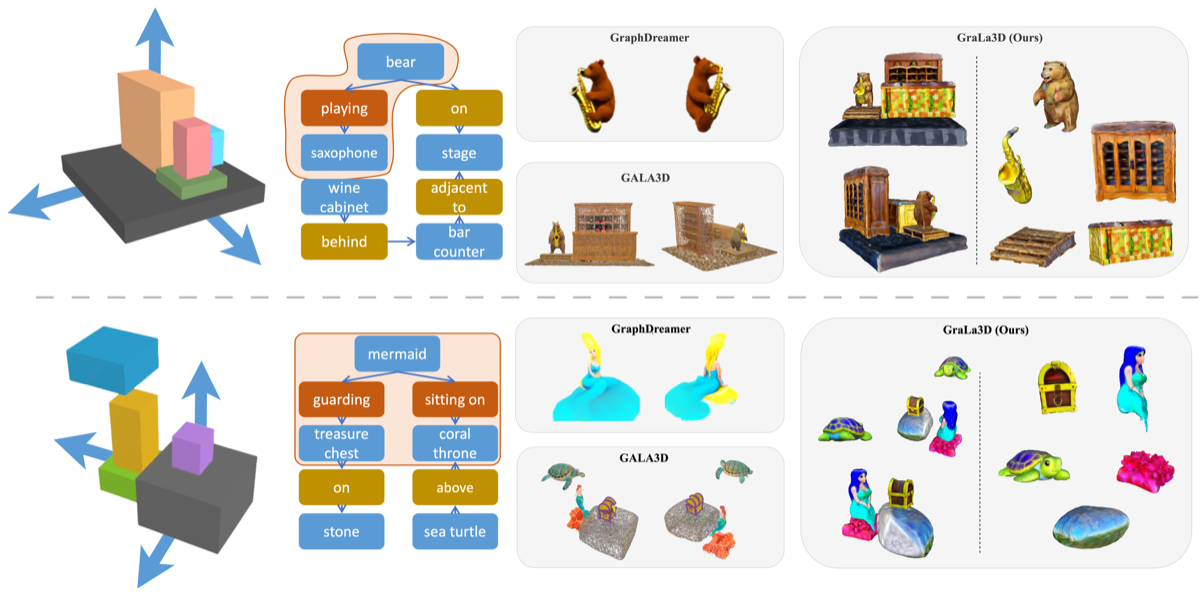}

    \vspace{-2mm}
    \caption{\textbf{Examples of text-to-3D generation with five objects.} Given the prompt of ``\textit{A mermaid sits on a coral throne, guarding a treasure chest on a stone while a sea turtle swims above her.''} and \textit{``A bear playing a saxophone stands on the stage, with a bar counter adjacent to it and a wooden wine cabinet filled with wine bottles behind the bar counter.''}, our GraLA3D is able to generate proper interactions of the mermaid-throne pair where the mermaid is sitting on the coral throne. Similarly, the bear-sax pair is also generated properly with the bear holding and playing the saxophone. Conversely, the mermaid generated by GALA3D is floating above the throne instead of sitting on it, and the bear's hand is not touching the saxophone at all. As for GraphDreamer, it fails to generate all five objects in both cases. We show their results by using only \textit{``mermaid sits on a coral throne''} and \textit{``a bear playing saxophone''} as input prompts.
}
    \label{fig:5obj_qualitative}
\end{figure*}

%% file: Figure/kitchen_qualitative.tex
\begin{figure*}[t!]
    \centering

        \includegraphics[width=1\linewidth]{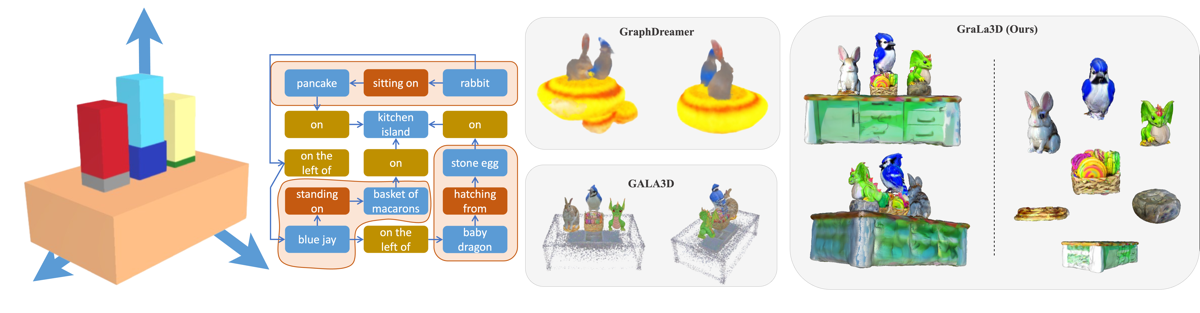}

    \vspace{-2mm}
\caption{\textbf{Example text-to-3D generation with seven objects.} Given the text prompt of ``\textit{A kitchen scene with a kitchen island. A rabbit sits on a stack of pancakes to the left of a blue jay standing on a large basket of rainbow macarons. On the right, a baby dragon hatches from a stone egg. The pancakes, large basket of rainbow macarons and stone egg are on top of the kitchen island
}'',  our GraLa3D is able to generate proper interaction such as the dragon-stone-kitchen island triplet, with the baby dragon on top of the stone egg, and both placed on the kitchen island. On the contrary, GALA3D failed to distinguish the dragon and the stone clearly. As for GraphDreamer, it fails to generate all seven objects in this case, and the maximum handled number of objects is 5 (i.e., the rabbit, the blue jay, the pancake, the basket of rainbow macarons, and the kitchen island).}
    \label{fig:kitchen_qualitative}
\end{figure*}

%% file: sec/4_experiments.tex
\section{Experiments}

\paragraph{Implementation Details.}
For the Scene Graph Composition stage, we conduct GPT4~\cite{openai2024gpt4technicalreport} to obtain the scene graph $G$, layout bounding box $B$, super-node $S$ and single-object nodes $O$. For the Node-to-3D Generation stage, we employ MVDream~\cite{Shi2023MVDreamMD} to calculate the ISM loss for 3000 iterations, with the rendered resolution progressively increasing from 128 to 512 throughout the process. Regarding the timestep sampling strategy, we follow DreamTime~\cite{huang2024dreamtime} and apply timestep sampling with monotonically non-increasing functions (for more implementation details, please refer to our supplementary material). In the 3D Scene Harmonization stage, we conduct 70 refinement iterations using the denoised images from ControlNet~\cite{zhang2023adding}, with depth as guidance for both individual objects and the global scene. 

\paragraph{Baseline methods.}
We compare our approach with two state-of-the-art methods for scene-level text-to-3D generation using their official implementations: GraphDreamer~\cite{Gao2023GraphDreamerC3} and GALA3D~\cite{Zhou2024GALA3DTT}. For a fair comparison, we provide GraphDreamer with the same scene graph input used in our method and apply identical layout bounding boxes for GALA3D. The training protocol for GraphDreamer includes 10,000 steps for a coarse stage followed by 10,000 steps for a fine stage. For GALA3D, we apply 15,000 iterations in alignment with the recommended configuration.

\subsection{Quantitative Results}

To quantitatively assess the alignment between the generated 3D models and the input text prompt, $y^g$, we use the CLIP Score as our primary metric. In this evaluation, we sample 200 random views per scene and compute the average CLIP score for each test case presented in Figure~\ref{fig:archi},~\ref{fig:4obj_qualitative},~\ref{fig:5obj_qualitative}, and~\ref{fig:kitchen_qualitative}. Our method achieves the highest CLIP scores across all tested scenes, demonstrating its superior ability to accurately decompose complex prompts and faithfully generate corresponding 3D scenes.

\subsection{Qualitative Results}
\label{sec:quali}
As illustrated in Figure~\ref{fig:archi},~\ref{fig:4obj_qualitative},~\ref{fig:5obj_qualitative}, and~\ref{fig:kitchen_qualitative}, our method, compared with both GraphDreamer and GALA3D, successfully generates a vivid scene with a clear layout and accurate interactions between objects. For GraphDreamer, although it is able to generate objects with interactions when the number of objects mentioned in the input text prompt is small enough (within four objects), it is not able to generate reasonable scenes for text prompts that incorporate over five objects with both spatial and interaction relations(Fig.~\ref{fig:archi},~\ref{fig:5obj_qualitative},~\ref{fig:kitchen_qualitative}), supporting our insight that the diffusion model struggles to guide spatial relationships or too many objects inside a scene effectively. The results shown for GraphDreamer in Figure~\ref{fig:4obj_qualitative} represent its most successful case, while in Figure~\ref{fig:archi},~\ref{fig:5obj_qualitative},~\ref{fig:kitchen_qualitative}, we reduce the number of objects in the text prompt when reproducing their results to show respectively reasonable results. GALA3D, on the other hand, generates each object separately without considering interactions between objects, leading to visual artifacts that are not consistent with the input prompts (e.g., the wizard gazing at the crystal ball, the rabbit eating a cake, or a mermaid sitting on a coral throne). Additionally, GALA3D omits adaptive density control and fixes the number of Gaussians for each object's 3DGS model, resulting in over-sparse representations for larger objects. Our approach successfully combines the strengths of both methods, proposing a novel solution for scene-level text-to-3D generation and extracting the 3D model in a readily available mesh format.

\subsection{Ablation Study}
Since the qualitative comparisons in Sect.~\ref{sec:quali} already show the necessity of our combination of layout and scene graph, we conduct additional ablation studies on the two losses introduced in our GraLa3D: the Localization loss and the Masked ISM loss.

\paragraph{Localization loss.}
\input{Figure/ablation_localization} 
To illustrate the misclassification of Gaussian blobs between two interactive related GS models (i.e., $\theta^S_1$ and $\theta^S_2$) during interaction loss optimization toward $\theta^S_1 \cup \theta^S_2$, we conducted an experiment where the localization loss, $\mathcal{L}_{\text{local}}$, was omitted. As shown in Figure \ref{fig:localization}, in the absence of the localization loss, parts of the instance (e.g., the astronaut's leg and the stone egg) are miscategorized into the other model (the horse and the dragon). Conversely, with an attention mechanism that constrains each instance $\theta^S_i$ to represent its specific object $v^S_i$, $\theta^S_1$ and $\theta^S_2$ can effectively disentangle from each other.

\paragraph{Masked ISM loss}
\input{Figure/maskedism}
To demonstrate the impact of the masked ISM loss $\mathcal{L}^{\text{ISM}}_{\text{mask}}$, we conduct an experiment for the prompt ``a horse''. As shown in Figure~\ref{fig:ism}, without the $\mathcal{L}^{\text{ISM}}_{\text{mask}}$, the 2D prior tends to generate structures that diverge from the intended layout. For example, the 3D model is guided toward a horse with its head raised, conflicting with the bounding box layout constraint and resulting in an incomplete head. In contrast, with $\mathcal{L}^{\text{ISM}}_{\text{mask}}$, the 3D model aligns with the desired layout, producing a horse with its head positioned forward as intended.

%% file: Figure/ablation_localization.tex
\begin{figure}[t]
    
    \includegraphics[width=\columnwidth]{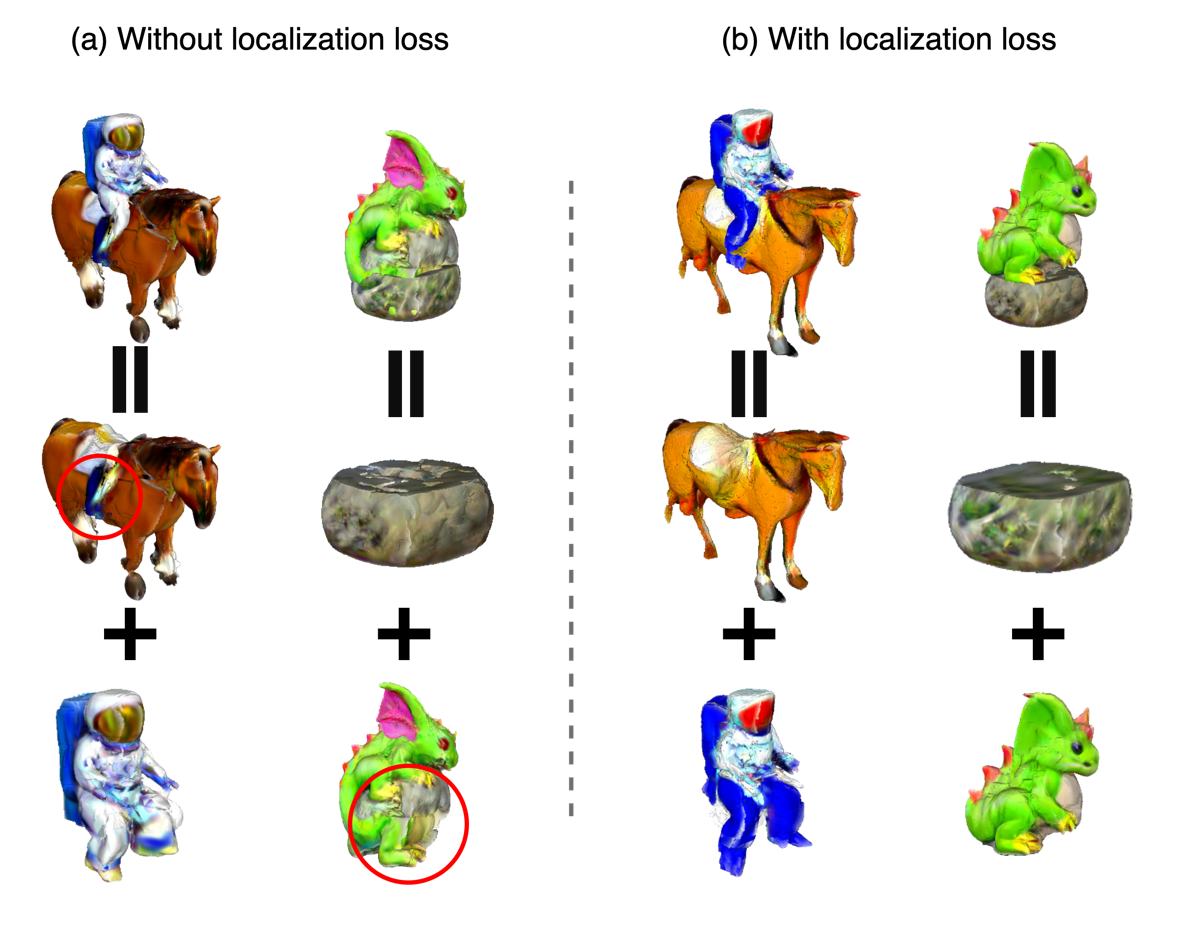}
    \vspace{-2mm}
    \caption{\textbf{Ablation study on the localization loss $\mathcal{L}_{\text{local}}$.} We show two examples of ``an astronaut riding a horse'' and ``a dragon hatches from a stone'' to show the effectiveness of $\mathcal{L}_{\text{local}}$
     Without $\mathcal{L}_{\text{local}}$, the composition of objects fails with parts of different instances entangling with each other (as highlighted in red circles).}
    \label{fig:localization}
\end{figure}

%% file: Figure/maskedism.tex
\begin{figure}[t]
    \centering
    \includegraphics[width=\columnwidth]{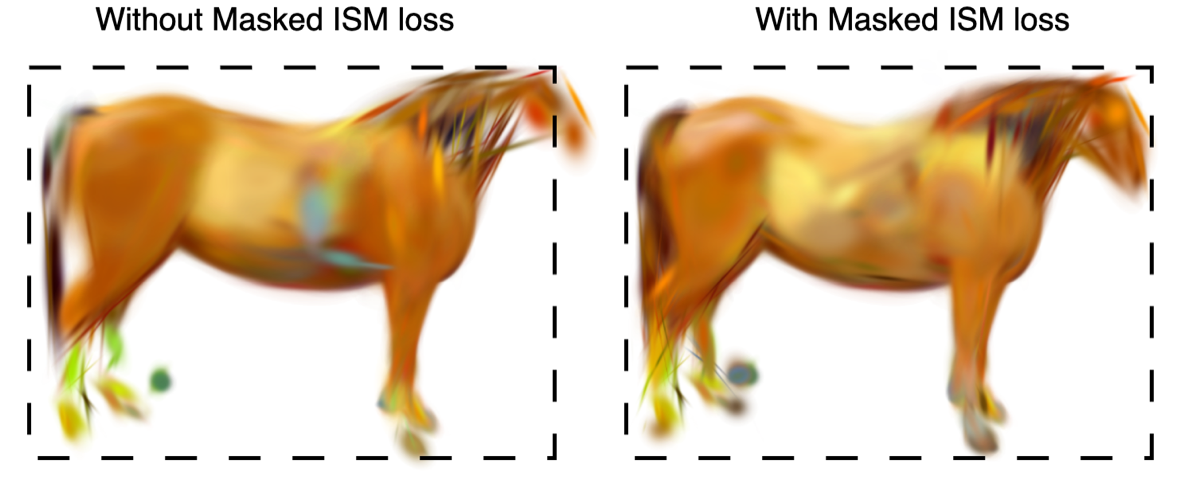}
    \vspace{-2mm}
    \caption{\textbf{Ablation study on $\mathcal{L}^{\text{msk}}_{\text{ISM}}$.} Without the masked ISM loss (replaced with the original ISM loss~\cite{Liang2023LucidDreamerTH}), the 2D diffusion prior~\cite{Shi2023MVDreamMD} produces an incomplete object (the horse head is cut off) to fit in the projected bounding box (denoted by the dashed rectangle). With $\mathcal{L}^{\text{mask}}_{\text{ISM}}$ as guidance, generation of an entire object fitting the bounding box can be achieved. }
    \label{fig:ism}
\end{figure}

%% file: sec/5_conclusion.tex
\section{Conclusion}
In this paper, we propose a Scene \textbf{Gra}ph and \textbf{La}yout Guided \textbf{3D} Scene Generation (GraLa3D) framework for complex 3D scene generation from text prompts. Our GraLa3D leverages the power of LLM to model complex 3D scenes via scene graphs accompanied with layout bounding boxes to manage both spatial and interactive object relations within a scene. By classifying nodes in the scene graph into single-object nodes and super-nodes based on different relation types, GraLa3D applies specialized Node-to-3D Generation strategies to ensure correct spatial properties and vivid interactions while preventing entanglements between objects that break the wholeness of individual objects. The final Scene Harmonization stage further aligns object styles, resulting in refined, visually consistent, and cohesive scenes. Comparative qualitative and quantitative evaluations with state-of-the-art methods highlight the strengths of our framework in generating complex 3D scenes while achieving close alignments with text prompts.

Since we perform mesh extraction from 3DGS to achieve global harmonization, the geometry of our generated 3D scene relies on the quality of this process. To overcome this limitation, studies on the 3DGS representation itself and its conversion to 3D mesh would be among future directions to improve the quality of our generation outputs.

%% file: sec/X_suppl.tex
\clearpage
\setcounter{page}{1}
\maketitlesupplementary

\appendix
\section{Additional implementation Details}

\subsection{LLM Prompting for Scene Graph and Layout Generation}

In the Scene Graph Composition stage (Section \textcolor{red}{3.2}), input text prompt $y^g$ is first converted into structured scene graph $G$ and bounding boxes $B$ using large language model (LLM). We now detail the use of LLM. We leverage the ChatGPT-o1-preview, a model equipped with chain-of-thought capabilities. To generate a scene graph, we provide a prompt specifying all the objects in scene, and then LLM will construct the scene graph for us. For example, to generate the farm scene with an astronaut riding a horse in Figure. \textcolor{red}{1}, our input prompt for the LLM is designed as:

\begin{quote}
"Generate a scene graph of a farm scene (farm is not included in scene graph) with nodes (objects) and edges (relations). The scene graph should conform to "An astronaut rides a brown horse on a farm, with a wooden fence in front, a barn beside them, and a haystack behind the barn." Use the minimum number of edges to construct a graph connecting all nodes."
\end{quote}

\begin{figure}[h!]
    \centering
    \includegraphics[width=\linewidth]{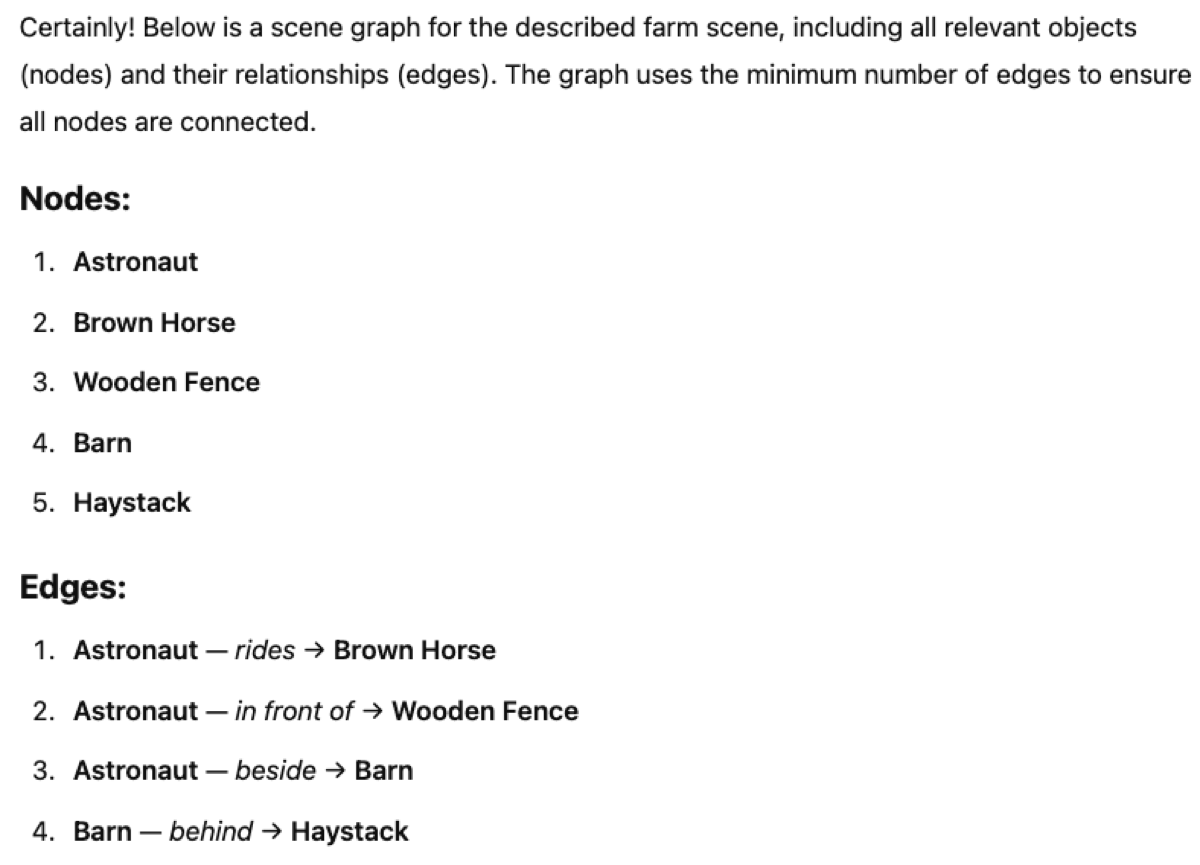}
    \caption{\textbf{ChatGPT for scene graph generation.} The result shows that ChatGPT is capable of composing the scene with main objects given.}
    \label{fig:gpt-scene-graph}
\end{figure}

As illustrated in Figure~\ref{fig:gpt-scene-graph}, the LLM constructs a scene graph. Then, as described in Figure \textcolor{red}{3}, the nodes \(V\) of graph \(G\) are decomposed into two groups: single-object nodes \(O\) and super-nodes \(S\), based on the spatial and interactive relations, which can also be classified using large language models (LLMs).

The generated scene graph serves as structured input for the LLM to predict the spatial layout of the scene. The prompt we use for generating the layout bounding boxes is: 
\begin{quote}
  \#ROLE: You are a expert in designing of realistic 3D computer graphic models. 
  
  \#CONTEXT: This is a scene with a scene graph. The  scene graph contains nodes: \{\textit{the nodes}\} and the edges: \{\textit{the edges}\}. With these information, you will know the textual description on objects and their relations. Lastly, currently floor is at y=0. follow the right-handed coordinate system
  
  \#OBJECTIVE: Predict the bounding box coordinates for all object nodes. Note that a bounding box must be a rectangular cuboid and is represented by 8 points in the Cartesian coordinate system with y=0 as physical ground. The unit of length of the coordinate system is 1 meter. Finally, please ensure the following criteria: 
  \begin{itemize}
      \item Relation Compliance: Strictly adhere to the object relationships provided in the following list: \{\textit{the edges}\}
  \item Non-Overlapping Constraint: Bounding boxes between \{\textit{the interactions}\} are encouraged to partially overlap but not entirely 
  \item Relative Sizing: The y axis is the height while the x, z are the width. Ensure that the relative sizes of the bounding boxes but the difference of size between boxes should not be too big.
 \item Physical Constraints: Remember that y=0 is the floor. All relations between bounding boxes must comply with physical constraints, such as gravity. 
  \item Real-World Size Reference: Consider the actual size of the object in reality when determining the dimensions of its bounding box and the size of different bounding box should not vary too much 
  \item Three-Dimensional Integrity: Each bounding box must have thickness in all x, y, and z dimensions; flat bounding boxes are not permitted. 
  \item Reference Usage: Utilize existing bounding boxes as reference points for positioning and sizing. 
  \end{itemize}
\end{quote}
\input{Figure/more_qualitative}
\input{Figure/huge_qualitative}

In this prompt, we replace each \{\textit{the nodes}\} and \{\textit{the edges}\} with the nodes and edges produced by the LLM-generated scene graph (as shown in Figure~\ref{fig:gpt-scene-graph}) and \{\textit{the interactions}\} are those edges classified as interactive. By using this prompt, the LLM generates a layout based on the scene graph.

\subsection{Cross-Attention Map Extraction for the Localization Loss}
As outlined in Section \textcolor{red}{3.3}, we derive the attention map $D^S$ from the rendered image $x^S$ associated with the prompt corresponding to each object described in the supernode $S$ as an instance-level pseudo ground truth for localization loss $\mathcal{L}_{\text{local}}$. This is achieved by employing the technique introduced in DAAM~\cite{tang-etal-2023-daam}. Specifically, DAAM aggregates the cross-attention maps across different timesteps, layers, and attention heads during the denoising process of diffusion models in text-to-image generation. This aggregation produces cross-attention maps that highlight the regions of the image that are attended by specific words (in our case, the object $v^S$) in the input text prompt. To extract the cross-attention map between the image $x^S$ and the prompt, we first iteratively add noises to the image $x^S$ to simulate the forward diffusion process up to half of the total inference steps (i.e., $25$, for a total $50$-step DDIM denoising process), following the best practices outlined in the original DAAM paper. We then denoise the noisy image from this intermediate timestep back to timestep $0$ with a pre-trained diffusion model. The aggregated cross-attention map during the denoising process is then treated as $D^S$ and is used for the calculation of $\mathcal{L}_{\text{local}}$.

\subsection{Hyperparameters and Optimization Strategy}

The optimization processes of each single-object node and each super-node are both 3000 iterations. For super-node generation, to prevent $\mathcal{L}_{\text{local}}$ from localizing meaningless regions while models at early steps of the optimization, we turn off $\mathcal{L}_{\text{local}}$ (by setting its weight to $0$) for the first 600 iterations. After the first 600 iterations. we turn on $\mathcal{L}_{\text{local}}$ and finish the optimization with our full objectives.


\section{Additional Experiments and Analysis}

\subsection{More Qualitative Evaluation}
We provide additional qualitative comparison and the result of our generation approach in Fig.~\ref{fig:6obj_qualitative}. Similar to the results in the main paper, our method demonstrates its capability to generate scenes with accurate interactions (e.g., the rabbit eating the cake) and spatial layouts (e.g., all the utensils properly arranged on the table). In contrast, GraphDreamer fails to produce the result when a specific scene layout is required, while GALA3D struggles with effectively modeling the interaction between the rabbit and the cake. 

Additionally, we provide the generation result of a scene with 15 objects and five supernodes in Fig.~\ref{fig:huge_qualitative}. This amount of objects is more than any case demonstrated in GALA3D or GraphDreamer, showing that our GraLa3D is not only capable of handling a large amount of objects but also generates proper interactions between them.

\subsection{User study}
\input{Table/user_study}
To thoroughly evaluate GraLa3D's ability to generate multi-object scenes, we designed a comparative survey involving six specific scenes (referenced in Figures \textcolor{red}{1}, \textcolor{red}{5}, \textcolor{red}{6}, \textcolor{red}{7}, and ~\ref{fig:6obj_qualitative}). The study compared the outputs of GraLa3D with those from baseline methods.

We invited 29 subjects to participate in this evaluation. Each rater reviewed all generated examples and, for each prompt, was asked to select the output that best matched the semantic content of the prompt. The results of their evaluations are summarized in Table~\ref{user}.

The findings demonstrate a clear preference for GraLa3D's outputs: 78.8\% of the raters selected results from GraLa3D as being more aligned with the prompts compared to those from the baseline methods. This outcome highlights GraLa3D's superior consistency and reliability in producing semantically accurate multi-object scenes.

\subsection{Limitations}
We now discuss the potential limitations of GraLa3D. First, we observed that in some cases, the layout fails to align with the intended interaction. For instance, given the prompt \textit{"a monkey holding a plate"}, the LLM sometimes generates two close but non-intersecting bounding boxes, which contradicts the interaction \textit{"holding"}. This small misalignment leads to generation a result that the plate is not attaching to the monkey's hand. In such cases, we further manually prompting the LLM to adjust the produced layout in the Layout Generation process. Future improvements could involve automated consistency-checking mechanisms to detect and resolve conflicts or fine-tuned LLMs capable of dynamically adjusting inputs, ensuring coherence without manual intervention.

Second, in the final stage of GraLa3D, global harmonization is performed after extracting the mesh from 3DGS. Following DreamGaussian~\cite{tang2024dreamgaussian}, we extract the occupancy field from 3DGS and apply the marching cubes algorithm to generate the mesh. However, sampling the occupancy of 3DGS leads to potential inaccuracies in mesh geometry. Future research could address this limitation by exploring constraints in 3DGS to improve the efficiency of 3D mesh conversion, as demonstrated in studies on efficient 3D mesh reconstruction~\cite{guedon2023sugar}, thereby enhancing output quality.

%% file: Figure/more_qualitative.tex
\begin{figure*}[t!]
    \centering

        \includegraphics[width=1\linewidth]{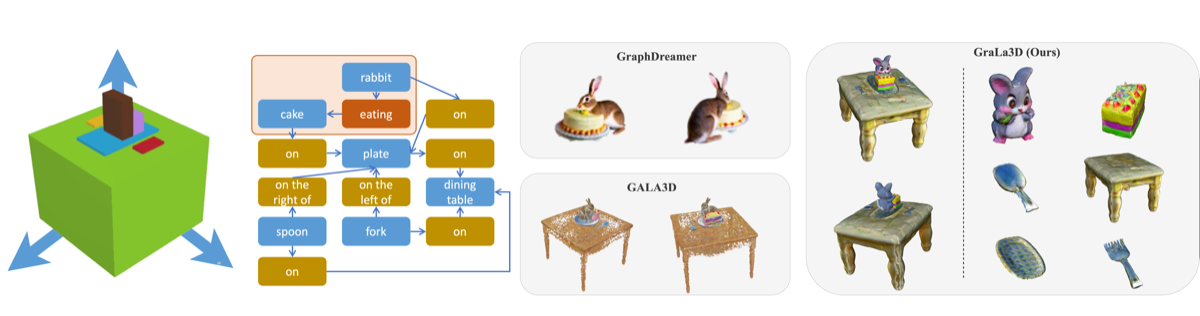}

    \vspace{-2mm}
    \caption{\textbf{Examples of text-to-3D generation with six objects.} Given the prompt of ``\textit{A rabbit is eating a cake on a plate. The plate, along with a spoon and fork, is on the table. The spoon is to the right of the plate, and the fork is to the left.''} Among the three methods, GraLa3D is the only one that correctly generates both the rabbit-eating-cake triplet and the precise positioning of the dining table, utensils, and the rabbit. This demonstrates that our method effectively combines the strengths of both scene graphs and layout-based approaches to generate scenes of greater complexity than those achieved by previous works.
}
    \label{fig:6obj_qualitative}
\end{figure*}

%% file: Figure/huge_qualitative.tex
\begin{figure*}[t!]
    \centering

        \includegraphics[width=1\linewidth]{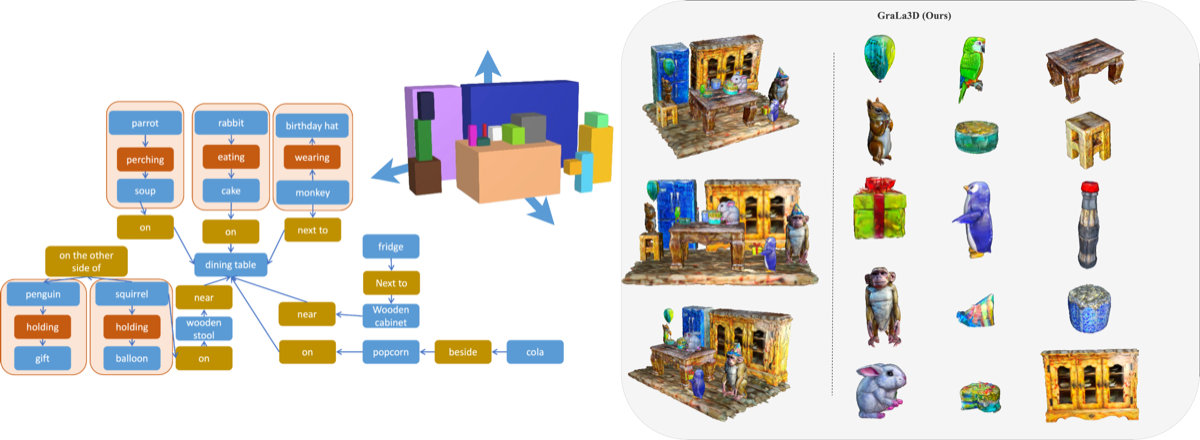}

    \vspace{-2mm}
    \caption{\textbf{Examples of text-to-3D generation with 15 objects.} Given the prompt of
``\textit{A lively scene in a dining room featuring a big dining table at the center. A monkey is next to the table, wearing a birthday hat. On the table, a rabbit is eating a birthday cake, and a parrot is perched on a bowl of soup placed beside the cake. On the other side of the table, a squirrel sits on a wooden stool holding a balloon, positioned near the table. Opposite the squirrel, a penguin holding a gift is located. The table is also set with a bowl of popcorn next to a bottle of cola. In the background, a wooden cabinet stands near the table, with a fridge next to the cabinet, completing the cozy dining setup.''}
GraLa3D demonstrates the ability to handle complex scenes effectively. Despite the presence of 15 distinct objects, our innovative Scene Graph Composition approach allows for a divide-and-conquer method that outperforms previous systems, enabling it to generate scenes of unprecedented complexity and detail.
}
    \label{fig:huge_qualitative}
\end{figure*}

%% file: Table/user_study.tex
\begin{table}[th]
\centering
\renewcommand{\arraystretch}{1.5} 
\resizebox{\linewidth}{!}{%
\begin{tabular}{c|ccc}
\toprule
& GraphDreamer~\cite{Gao2023GraphDreamerC3} & GALA3D~\cite{Zhou2024GALA3DTT} & \cellcolor[gray]{0.9}GraLa3D (Ours) \\ \midrule
Farm case (Fig. \textcolor{red}{1})     & 0.0         & 10.3     & \cellcolor[gray]{0.9}89.7     \\
Wizard case (Fig. \textcolor{red}{5})  & 6.9         & 3.4      & \cellcolor[gray]{0.9}89.7     \\
Kitchen case (Fig. \textcolor{red}{7}) & 3.4         & 6.9      & \cellcolor[gray]{0.9}89.7     \\
Mermaid case (Fig. \textcolor{red}{6}) & 0.0         & 37.9     & \cellcolor[gray]{0.9}62.1     \\
Bear case  (Fig. \textcolor{red}{6})   & 0.0         & 37.9     & \cellcolor[gray]{0.9}62.1     \\
Rabbit case (Fig.~\ref{fig:6obj_qualitative})  & 0.0         & 20.7     & \cellcolor[gray]{0.9}79.3     \\ \midrule
Avg.          & 1.7         & 19.5     & \cellcolor[gray]{0.9}78.8     \\
\bottomrule
\end{tabular}}
\caption{\textbf{Result for user study (unit: \%).} In each case, our generated result is selected as the one most semantically aligned with the prompt describing the scene, demonstrating the capability of our method in handling complex scenes.
}
\label{user}
\end{table}